\documentclass[
]{ceurart}

\begin{document}

%%
%% Rights management information.
%% CC-BY is default license.
\copyrightyear{2020}
\copyrightclause{Copyright for this paper by its authors.\\
  Use permitted under Creative Commons License Attribution 4.0
  International (CC BY 4.0).}

%%
%% This command is for the conference information
\conference{FIRE 2020: Forum for Information Retrieval Evaluation, December 16-20, 2020, Hyderabad, India}

%%
%% The "title" command
\title{CMSAOne@Dravidian-CodeMix-FIRE2020: A Meta Embedding and Transformer model for Code-Mixed Sentiment Analysis on Social Media Text}

%%
%% The "author" command and its associated commands are used to define
%% the authors and their affiliations.
\author[1]{Suman Dowlagar}[%
orcid=0000-0001-8336-195X,
]
\ead[1]{suman.dowlagar@research.iiit.ac.in}
\address[1]{International Institute of Information Technology - Hyderabad (IIIT-Hyderabad),
  Gachibowli, Hyderabad, Telangana, India, 500032}

\author[1]{Radhika Mamidi}
\ead[2]{radhika.mamidi@iiit.ac.in}

%%
%% The abstract is a short summary of the work to be presented in the
%% article.
\begin{abstract}
  Code-mixing(CM) is a frequently observed phenomenon that uses multiple languages in an utterance or sentence. CM is mostly practiced on various social media platforms and in informal conversations. Sentiment analysis (SA) is a fundamental step in NLP and is well studied in the monolingual text. Code-mixing adds a challenge to sentiment analysis due to its non-standard representations. This paper proposes a meta embedding with a transformer method for sentiment analysis on the Dravidian code-mixed dataset. In our method, we used meta embeddings to capture rich text representations. We used the proposed method for the Task: ``Sentiment Analysis for Dravidian Languages in Code-Mixed Text'', and it achieved an F1 score of $0.58$ and $0.66$ for the given Dravidian code mixed data sets. The code is provided in the Github \url{https://github.com/suman101112/fire-2020-Dravidian-CodeMix}.
\end{abstract}

%%
%% Keywords. The author(s) should pick words that accurately describe
%% the work being presented. Separate the keywords with commas.

%%
%% This command processes the author and affiliation and title
%% information and builds the first part of the formatted document.
\maketitle

\section{Introduction}

Code-mixing(CM) of text is prevalent among social media users, where words of multiple languages are used in the sentence. Code-mixing occurs when conversant uses both languages together to the extent that they change from one language to another in the course of a single utterance \cite{wardhaugh2011introduction}. The computational modeling of code-mixed text is challenging due to the linguistic complexity, nature of mixing, the presence of non-standard variations in spellings, grammar, and transliteration \cite{bali2014borrowing}. Because of such non-standard variations, CM poses several unseen difficulties in fundamental fields of natural language processing (NLP) tasks such as language identification, part-of-speech tagging, shallow parsing, Natural language understanding, sentiment analysis. 

\citet{gysels1992french} defined the Code-mixing as ``the embedding of linguistic units of one language into an utterance of another language''. Code-mixing is broadly classified into two types, intra-sentential and inter-sentential. Intra-sentential code-mixing happens after every few words. Whereas, in inter-sentential code-mixing, one part of the sentence consists of Hindi words, and the other part is entirely English. The code-mixing helps people to express their emotions or opinions emphatically, thus leading to a phenomenal increase of use in code-mixed messages on social media platforms. With the increase in code-mixed data, the analysis of CMSM text has become an essential research challenge from the perspectives of both Natural Language Processing (NLP) and Information Retrieval (IR) communities. 

There have been some research works in this direction, such as GLUECoS, an evaluation benchmark in code-mixed text \cite{khanuja2020gluecos},  automatic word-level language identification for CMSM text \cite{barman2014code,gella2014ye}, parsing pipeline for Hindi-English CMSM text \cite{sharma2016shallow,nelakuditi2016part}, and
POS tagging for CMSM text \cite{vyas2014pos}.

To encourage research on code-mixing, the NLP community organizes several tasks and workshops such as Task9: SentiMix, SemEval 2020\footnote{https://competitions.codalab.org/competitions/20654}, and 4th Workshop on Computational Approaches for Linguistic Code-Switching\footnote{https://www.aclweb.org/portal/content/fourth-workshop-computational-approaches-linguistic-code-switching}. Similarly, the FIRE 2020's Dravidian-CodeMix task\footnote{https://dravidian-codemix.github.io/2020/} was devoted to code-mixed sentiment analysis on Tamil and Malayalam languages. This task aims to classify the given CM youtube comments into one of the five predefined categories: positive, negative, mixed\_feelings,  not\_\textless language\textgreater\footnote{the language might be Tamil and Malayalam}, unknown\_state.

In this paper, we present a meta-embedding with a transformer model for Dravidian Code-Mixed Sentiment Analysis. Our work is similar to the meta embedding approach used for named entity recognition on code-mixed text \cite{priyadharshini2020named}.

The paper is organized as follows. Section 2 provides related work on code-mixed sentiment analysis. Section 3 describes the proposed work. Section 4 presents the experimental setup and the performance of the model. Section 5 concludes our work.

\section{Related Work}

Sentiment analysis is one of the essential tasks in the field of NLP. Sentiment analysis is the process of understanding the polarity of the sentence. Sentiment analysis helps to attain the public's attitude and mood, which can help us gather insightful information to make future decisions on large datasets \cite{liu2020sentiment}. Initially, sentiment analysis was used on government campaigns and news articles \cite{tayal2017sentiment,godbole2007large}. Recently, due to social media prevalence, the research turned towards capturing the sentiment on social media texts in code-mixing scenarios \cite{patwa2020semeval}. 

The earlier approaches used syntactic rules and lexicons to extract features followed by traditional machine learning classifiers for sentiment analysis on code-mixed text. The process of rule extraction and defining lexicons is a time consuming, laborious process, and is domain-dependent. The recent work in the field of CMSA uses embeddings with deep learning and traditional classifiers \cite{mishra2018code}. The paper \cite{prabhu2016towards} used sub-word information for sentiment analysis on code-mixed text. The recent SentiMix 2020 task used BERT-like models and ensemble methods to capture the code-mixed texts' sentiment \cite{patwa2020semeval}. We used meta embeddings with state of the art transformer model for this task.

\section{Proposed Model}

This section presents our proposed code-mixed sentiment analysis framework. It has three main components: a sub-word level tokenizer, a text representation layer, and a transformer model.

\subsection{Sub-word Level Tokenizer}
To deal with the non-standard variations in spellings, we used the SentencePiece \cite{kudo2018sentencepiece}. SentencePiece is an unsupervised text tokenizer and de-tokenizer mainly used for neural network models. SentencePiece treats the sentences just as sequences of Unicode characters. It implements subword units by using byte-pair-encoding (BPE) \cite{sennrich2015neural} and unigram language model \cite{kudo2018subword} . The byte pair encoding initializes the vocabulary to every character present in the corpus and progressively learn a given number of merge rules. The unigram language model trains the model with multiple subword segmentations probabilistically sampled during training.

\subsection{Text Representation Layer}

Pre-trained embedding models do not perform well on the code-mixed corpus as they consider all the code-mixed words as OOV words \cite{pratapa2018word}. Thus, we have to train word representations from the code-mixed corpus. Given the complexity of the code-mixed data, it is not easy to determine which embedding model to be used for better performance. Hence, we chose the combination of fastText \cite{bojanowski2017enriching}, ELMO \cite{peters2018deep}, and TF-IDF \cite{aizawa2003information} embeddings. fastText captures efficient text representations and local dependencies at the word and sub-word level. ELMO captures contextual representations at the sentence level. TF-IDF captures the distribution of the words in the corpus. The use of TF-IDF for sentiment analysis helps in extracting a better correlation between words and their polarity. All these diverse text representations, when combined, proved beneficial in obtaining better embeddings for the downstream tasks.

\subsection{Transformer model}

From \cite{patwa2020semeval}, For the task of sentiment analysis, we saw that the attention mechanism works better in deciding which part of the sentence is essential for capturing the sentiment. Thus we chose the transformer model for our code-mixed sentiment analysis task. As the data is a classification type, we used only the encoder side from the Transformer. 

The encoder encodes the entire source sentence into a sequence of context vectors. First, the tokens are passed through a standard embedding layer, and the positional embeddings are concatenated with each source sequence. The embeddings are then passed through a series of encoder layers to get an encoded sequence.

The encoder layers is an essential module where all the processing of the input sequence happens. We first pass the source sentence and its mask into the multi-head attention layer, then perform dropout, apply a residual connection, and pass it through a normalization layer. We later pass it through a position-wise feedforward layer and then, again, apply dropout, a residual connection, and layer normalization to get encoded output sequence. The output of this layer is fed into the next encoder layer. 

The Transformer model uses scaled dot-product attention given in equation \ref{eq1}, where the query $Q$ and key $K$ are combined by taking the dot product between them, then applying the softmax operation and scaled by a scaling factor $d_k$ then multiplied by the value $V$. Attention is a critical unit in the Transformer model as it helps in deciding which parts of the sequence are important.

\begin{equation}
 Attention(Q,K,V) = Softmax(\frac{ QK^{T}}{ \sqrt[]{d_k}})V
 \label{eq1}
\end{equation}

The other main block inside the encoder layer is the position-wise feedforward layer. The input is transformed from hid\_dim to pf\_dim, where pf\_dim is usually a lot larger than hid\_dim. The ReLU activation function and dropout are applied before it is transformed back into a hid\_dim representation. The intuition borrows from infinitely wide neural networks.  The wide neural network grants more approximation power and helps to optimize the model faster.

\subsection{Our Approach}

\begin{figure}
  \centering
  \includegraphics[width=\linewidth]{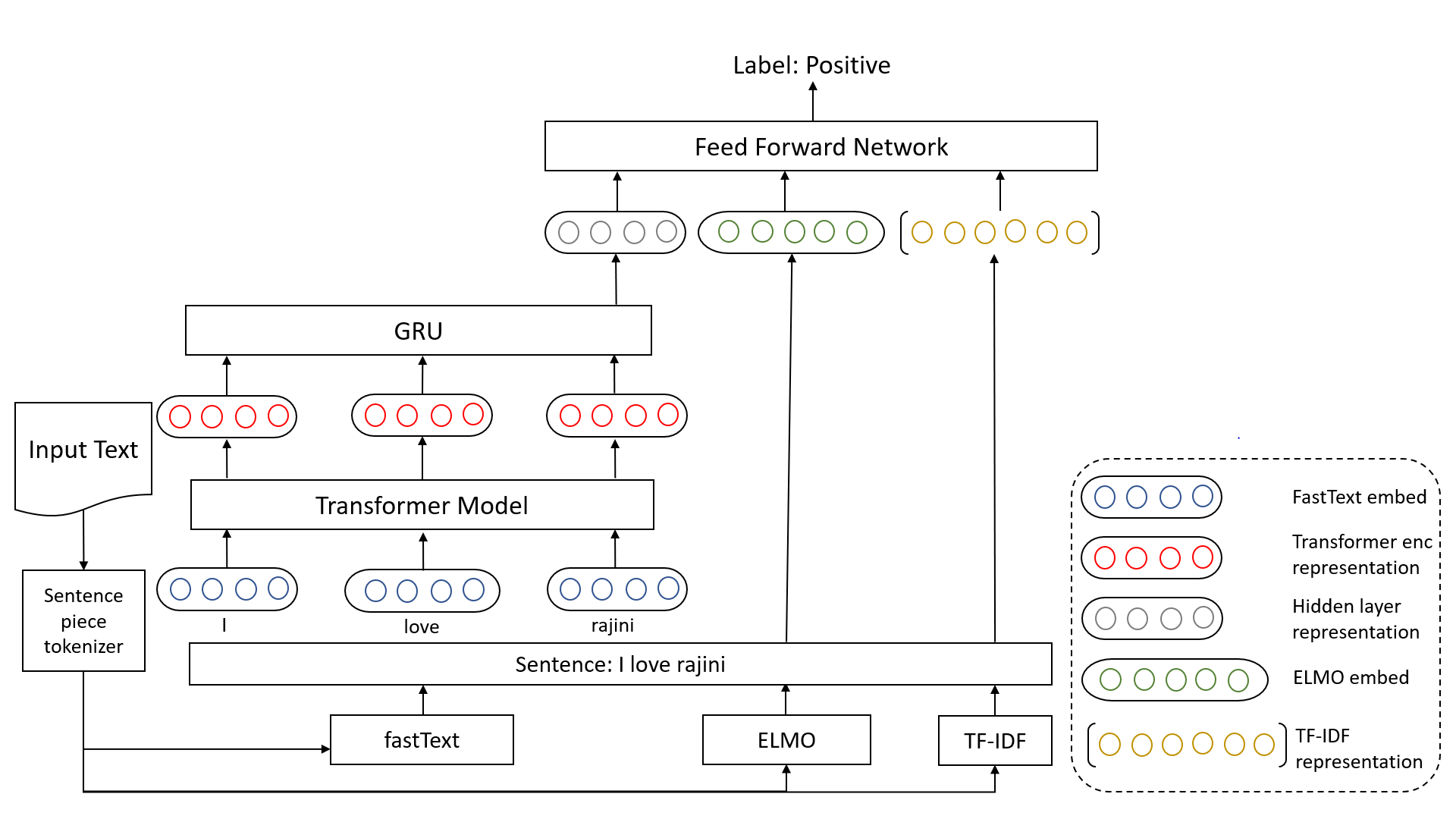}
  \caption{Meta Embedding with transformer and GRU model}
\end{figure}

Initially, we tokenized the sentence using the SentencePiece model.
After tokenization, we extracted local dependencies between embeddings at the subword level using the fastText model. The fastText model gave embeddings at the word level. We then applied the transformer model to obtain the encoded representations. We got the encoded representations at the word level. A GRU unit is used to get the encoded representation of all the words. We considered the representation of the last hidden layer of the GRU as the final encoded representation. We then obtained the ELMO contextual and TF-IDF representations at the sentence level.  We concatenated the representations of the last hidden GRU layer, ELMO, and TF-IDF giving us the meta-embeddings. The meta embeddings are then passed to the output feed-forward network to predict the polarity of the sentence.

\section{Experimental Setup}

\subsection{Data}
For Dravidian code-mixed sentiment analysis, we used the dataset provided by the organizers of Dravidian Code-mixed FIRE-2020. The training dataset consists of 15,744 Tamil CM and 6,739  Malayalam CM youtube video comments. The details of the dataset and the initial benchmarks on the corpus are given in \cite{dravidiansentiment-ceur,dravidiansentiment-acm,chakravarthi-etal-2020-sentiment,chakravarthi-etal-2020-corpus,chakravarthi2020leveraging}

\subsection{Hyperparameters}

\paragraph{For Embedding models}
Embeddings play a vital role in improving the model's performance. As mentioned above, we used fastText and ELMO embeddings. The dimensionality was set to 300 in-case of fastText embeddings. The embeddings are trained on training data using the parameters: learning rate = 0.05, context window = 5, epochs = 20. The ELMO model is obtained from tensorflow\_hub\footnote{https://tfhub.dev/google/elmo/2}, and the pre-set dimensionality of 1024 is used.

\paragraph{For Transformer and GRU model}
After evaluating the model performance on the validation data, the optimal values of the hyper-parameters were set. We used the following list of hyper-parameters: learning rate = 0.0005, transformer encoder layer = 1, dropout rate = 0.1, optimizer = Adam, loss function = Cross-Entropy Loss, and batch size = 32, point wise feed forward dimension (pf\_dim) = 2048.
\subsection{Performance}

\begin{table}
  \caption{Accuracy and weighted F1 score on Tamil Code-Mixed Text}
  \label{tab:results1}
  \begin{tabular}{lccc}
    \toprule
    Method & Accuracy & weighted F1\\
    \midrule
    %TF-IDF + SVM & 0.66 & 0.53\\
    %ELMO + TF-IDF + SVM & 0.64 & 0.52\\
    Fine Tuned BERT & 0.65 & 0.53\\
    fastText + Tranformer& 0.66 & 0.57\\
    fastText + ELMO + Transformer& 0.66 & 0.57\\
    \textbf{fastText + ELMO + TF-IDF + Transformer}& \textbf{0.67} & \textbf{0.58} \\
  \bottomrule
\end{tabular}
\end{table}

\begin{table}
  \caption{Accuracy and weighted F1 score on Malayalam Code-Mixed Text}
  \label{tab:results2}
  \begin{tabular}{lccc}
    \toprule
    Method & Accuracy & weighted F1\\
    \midrule
    %TF-IDF + SVM & 0.67 & 0.63\\
    %ELMO + TF-IDF + SVM  & 0.67 & 0.63 \\
    Fine Tuned BERT & 0.51 & 0.46\\
    fastText + Tranformer& 0.47 & 0.45\\
    fastText + ELMO + Transformer& 0.50 & 0.47\\
    \textbf{fastText + ELMO + TF-IDF + Transformer}& \textbf{0.67}& \textbf{0.66}\\
  \bottomrule
\end{tabular}
\end{table}

We evaluated the performance of the method using weighted F1. The model performed well in classifying positive and not-{language} comments. The results are given in table \ref{tab:results1} and \ref{tab:results2} The positive comments had a lot of corpora to train. It made the classification of positive comments an easier task. The not\_Malayalam and not\_Tamil tweets had another language words in the data, as these language words had higher TF-IDF scores w.r.t the non-{language} label, their classification was straight-forward. We observed that the system could not identify the sentiment when sarcasm is used in the negative polarity comments. The words in the sarcasm are similar to those of positive comments. It made the sentiment analysis a difficult task. 

Mixed feelings had both positive and negative sentences. As the classifier was trained on a lot of positive corpora, it could not deduce the negative polarity sentences with sarcasm and irony imbibed in them. Thus the classifier labeled them as positive. It affected the performance of the classifier. More training data could help resolve such issues.

\section{Conclusion}

This paper describes the approach we proposed for the Dravidian Code-Mixed FIRE-2020 task: Sentiment Analysis for Davidian Languages in Code-Mixed Text. We proposed meta embeddings with the transformer and GRU model for the sentiment analysis of Dravidian code mixed data set given in the shared task. Our model obtained 0.58 and 0.66 average-F1 for Tamil and Malayalam code-mixed datasets, respectively. We observed that the proposed model did a good job distinguishing positive and not\_Malayalam and not\_Tamil youtube comments.  For future work, we will explore our model's performance with larger corpora. As we observed sarcasm and irony in negative polarity sentences, we feel that it would be interesting to focus on techniques to detect irony and sarcasm in a code-mixed scenario.

%%
%% Define the bibliography file to be used
\bibliography{sample-ceur}

%%
%% If your work has an appendix, this is the place to put it.
\appendix

\end{document}